\documentclass[sigconf]{acmart}
\usepackage{listings}

\renewcommand\footnotetextcopyrightpermission[1]{} 

\AtBeginDocument{%
  \providecommand\BibTeX{{%
    \normalfont B\kern-0.5em{\scshape i\kern-0.25em b}\kern-0.8em\TeX}}}

\graphicspath{{./images/}} 
\begin{document}

\title{A Topological Improvement of the Overall Performance of Sparse Evolutionary Training: Motif-Based Structural Optimization of Sparse MLPs Project}

\author{
  Xiaotian Chen$^{1}$, 
  Hongyun Liu$^{1}$, 
  Seyed Sahand Mohammadi Ziabari$^{1,2}$
}

\affiliation{%
  \institution{$^{1}$Informatics Institute, University of Amsterdam}
  \streetaddress{1098XH Science Park}
  \city{Amsterdam}
  \country{The Netherlands}
}

\affiliation{%
  \institution{$^{2}$Department of Computer Science and Technology, SUNY Empire State University}
  \city{Saratoga Springs, NY}
  \country{USA}
}

\email{chenxiaotian097@gmail.com, h.liu@uva.nl, sahand.ziabari@sunyempire.edu}

\begin{abstract}
Deep Neural Networks (DNNs) have been proven to be exceptionally effective and have been applied across diverse domains within deep learning. However, as DNN models increase in complexity, the demand for reduced computational costs and memory overheads has become increasingly urgent. Sparsity has emerged as a leading approach in this area. The robustness of sparse Multi-layer Perceptrons (MLPs) for supervised feature selection, along with the application of Sparse Evolutionary Training (SET), illustrates the feasibility of reducing computational costs without compromising accuracy. Moreover, it is believed that the SET algorithm can still be improved through a structural optimization method called \textit{motif-based optimization}, with potential efficiency gains exceeding 40\% and a performance decline of under 4\%. This research investigates whether the structural optimization of Sparse Evolutionary Training applied to Multi-layer Perceptrons (SET-MLP) can enhance performance and to what extent this improvement can be achieved.
\end{abstract}


\fancyhead{}
\acmConference[]{}{}{}
\maketitle

\section{Introduction}
\label{sec:introduction}
The emergence of neural networks has facilitated the development of artificial intelligence (AI), defined as the ability of machines to simulate human cognitive processes. With the advancement of neural networks, the tasks they address have become increasingly complex, often involving the handling of high-dimensional data to satisfy specific requirements. To enhance performance, deep neural networks have evolved to more accurately mimic human brain functions, leading to substantial increases in computational cost and training time \cite{ardakani_sparsely-connected_2017}
. Typically,DNNs have many layers with fully-connected neurons, which contain most of the network parameters (i.e. the weighted connections), leading to a quadratic number of connections with respect to their number of neurons\cite{bellec_deep_2018}.

To address this issue, the concept of sparse connected Multi-layer Perceptrons with evolutionary training was introduced. This algorithm, when compared to fully connected DNNs, can substantially reduce computational cost on a large scale. Moreover, with feature extraction, sparsely connected DNNs can maintain performance comparable to that of fully connected models.

However, this approach still demands considerable computational resources and time, which remains a limitation. Furthermore, the introduction of motif-based DNNs, which can retrain neurons using small structural groups (e.g., groups of three neurons), suggested the potential to surpass the performance of sparse connected DNNs and significantly enhance overall network efficiency. This paper aims to analyze and test motif-based DNNs, comparing their performance against benchmark models.

To provide a deeper understanding, the following sections will delve into the foundational aspects of these approaches.

 As mentioned before, traditional neural networks are usually densely connected, meaning that each neuron is connected to every other neuron in the previous layer, resulting in a large number of parameters. Unlike normal DNNs models, SET helps introduce sparsity, reduce redundant parameters in the network, and improve computational efficiency. Through the evolutionary algorithm, SET can gradually optimize the weights so that many connections become irrelevant or zero\cite{bellec_deep_2018}. Therefore, SET is applied to improve training efficiency by optimizing the sparse structure of the model and reducing redundant parameters, which eventually can end in a reduction of the computational cost\cite{mocanu_scalable_2018}.

To further improve the performance of the Deep Neural Network, feature engineering is considered as a critical step in the development of machine learning models, involving the selection, extraction, and transformation of raw data into meaningful features that enhance model performance \cite{zhang_topology_2022}. By enforcing sparsity in the neural network, SET effectively prunes less important connections, thereby implicitly selecting the most relevant features. As the evolutionary algorithm optimizes the network, connections that contribute insignificantly to the model's performance are gradually set to zero, allowing the network to focus on the most informative features. If feature selection can be applied in this process, with some important features selected and remaining features dropped, the complexity of the network would be largely decreased and the quantified features would keep the original accuracy. Consequently, SET feature selection results in a streamlined model that is both computationally efficient and more accurate, leading to better overall performance\cite{mocanu_scalable_2018}, and Neil Kichler has demonstrated the effectiveness and robustness of this algorithm in his studies, further validating its practical application and benefits\cite{kichler_robustness_2021}.

Network motifs are significant, recurring patterns of connections within complex networks. They reveal fundamental structural and functional insights in systems like gene regulation, ecological food webs, neural networks, and engineering designs. By comparing the occurrence of these motifs in real versus randomized networks, researchers can identify key patterns that help to understand and optimize various natural and engineered systems.

\begin{figure}[h] 
    \centering
    \includegraphics[width=0.8\linewidth]{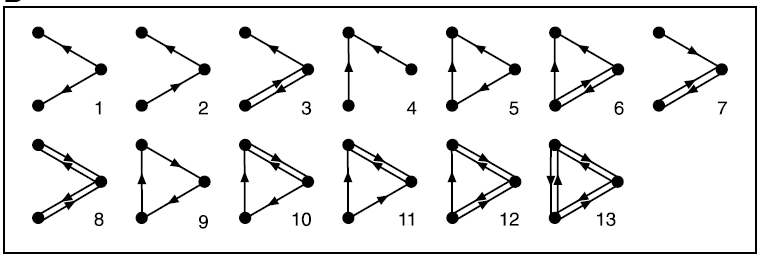}
    \caption{Motif-based Structural Sparse MLPs \cite{milo_network_2002}  }
    \label{fig:1}
\end{figure}

As mentioned before, the SET updates new random weights when the weights of the connections are negative or insignificant (close to or equal to zero) which to some extent lead to some computation burden\cite{altman_introduction_1992}. Based on the concept of motif and SET, a structurally sparse MLPs is proposed. The motif-based structural optimization gave an idea of renewing the weights by establishing a topology which can largely improve the efficiency (shown in Figure \ref{fig:1})\cite{liu_topological_2020,milo_network_2002}.

The key research question is posed in this paper is:   

\begin{center}
 \textbf{\textit{To what extent can the efficiency and accuracy of sparse MLPs get improved by optimizing the structure of the Sparse MLPs and fine-tuning of the network parameters?}}
\end{center}

\label{subsec:RQ}

\section{Related Work}
\label{sec:related_work}

Sparse MLP models have demonstrated significant potential in reducing computational costs (e.g., hardware and computation time) while enhancing accuracy through feature extraction and sparse training. This research uses the work of Mocanu et al.~\cite{mocanu_scalable_2018} as a benchmark model for comparison. This section reviews the historical development of sparse neural networks. Subsequently, the key idea and algorithm of SET will be discussed. Lastly, the basic idea of structural optimization for sparse MLPs will be introduced.

Y.~LeCun et al.~\cite{lecun_optimal_1989} introduced the concept of network pruning in the paper \textit{Optimal Brain Damage}. This approach computed the contribution of each connection to overall network error and selectively removed less important nodes. Utilizing second-order derivatives, this method effectively reduced model complexity while preserving performance, laying a theoretical foundation for later pruning techniques.

Building on this, B.~Hassibi et al.~\cite{hassibi_optimal_1993} proposed the \textit{Optimal Brain Surgeon} method in 1993, which also used second-order derivatives but provided a more precise pruning mechanism by considering the Hessian matrix. This refinement significantly improved pruning efficiency.

In 2016, Han et al.~\cite{han_deep_2016} introduced \textit{Deep Compression}, which combined pruning, quantization, and Huffman encoding. This three-step method substantially reduced storage and computational requirements while maintaining model accuracy. In addition to standard sparse training and retraining, the inclusion of Huffman encoding emphasized the advantage of integrating multiple optimization methods.

\begin{figure}[H]
    \centering
    \includegraphics[width=0.4\linewidth]{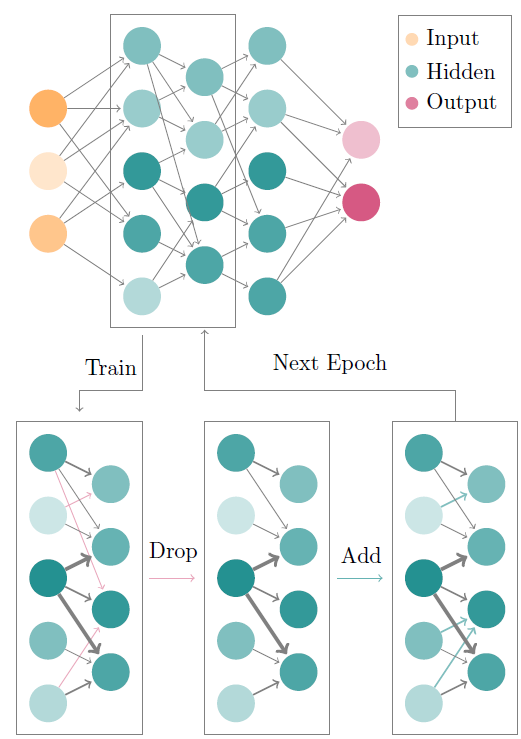}
    \caption{Process of training, pruning, and retraining in SET~\cite{kichler_robustness_2021}}
    \label{fig:2}
\end{figure}

In 2018, Mocanu et al.~\cite{mocanu_scalable_2018} proposed \textit{Sparse Evolutionary Training (SET)}. This approach used Erdős–Rényi graph initialization to create an initial sparse network, selectively adding and removing connections based on performance deviations. SET maintains a high ratio of zero-valued weights while optimizing accuracy. The SET training pipeline is illustrated in Figure~\ref{fig:2} and is used as the benchmark in this research.

Further advancing sparse model training, Frankle et al.~\cite{mostafa_parameter_2018} introduced \textit{Dynamic Sparse Reparameterization}, which adaptively adjusts network sparsity during training to maintain performance while improving efficiency. This approach stood out by dynamically reoptimizing structure, resulting in more effective training.

Building upon Mocanu’s SET, Kichler~\cite{kichler_robustness_2021} combined supervised feature selection with sparse multi-layer training. The study showed that even with significant feature pruning, the network retained performance comparable to fully connected models.

\begin{figure}[H]
    \centering
    \includegraphics[width=\linewidth]{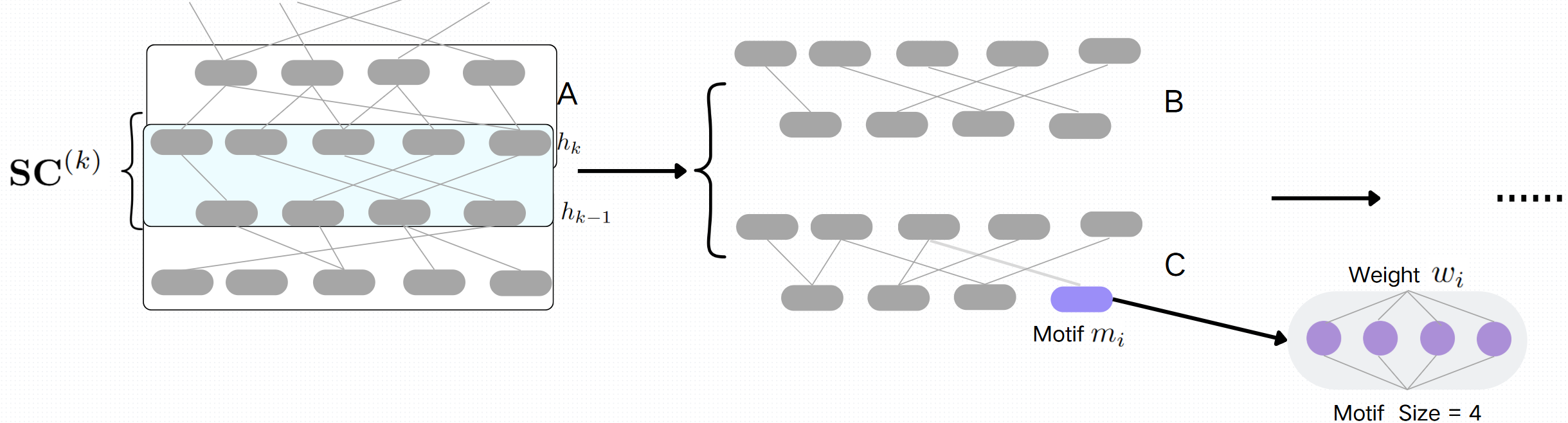}
    \caption{Concept of motif-based SET training}
    \label{fig:different}
\end{figure}

The motif-based concept refers to a specific type of structural topology or network pattern, as illustrated in Figure~\ref{fig:different}. In this visualization, the left-side DNN is trained and retrained at the individual node level~\cite{yin_dimensionality_2018}, while the motif-based model on the right applies training to small groups of nodes~\cite{bin_topology_2019}. By grouping nodes into motifs and assigning shared weights, the model improves training efficiency while maintaining accuracy.

\section{Methodology}
\label{sec:methodology}

To address the research questions related to the motif-based structural optimization of sparsely connected neural networks, this section provides a detailed illustration of the proposed approaches. First, it discusses the topological optimization method. Then, the training process is explained. Finally, the evolution mechanism for the motif-based SET model is described.

The core principle of motif-based structural optimization in SET involves assigning weights between neurons based on motifs during each training process, followed by distributing these weights to individual nodes.

\textbf{Listing 1: General Motif-Based Sparse Neural Network Process}
\begin{flushleft}
Initialize network with motif\_size: \\
\quad For each layer: \\
\quad\quad Initialize sparse weights and biases with motifs. \\[0.5ex]
Forward pass: \\
\quad For each hidden layer: \\
\quad\quad Process each motif with weights and biases. \\
\quad\quad Apply activation function. \\[0.5ex]
Backward pass: \\
\quad For each hidden layer in reverse: \\
\quad\quad Calculate delta. \\
\quad\quad Update weights and biases for each motif.
\end{flushleft}
\vspace{-1.52em} 
\subsection{Network Construction}

The general idea of using structural optimization based on motifs is to group nodes of a certain size and train them together. Unlike simply reducing the number of neurons, each node in the motif-based optimized network participates in both training and retraining. The key difference lies in the process of assigning new weights to nodes, which is conducted according to a specific topology, thereby enhancing the network's efficiency~\cite{hayase_mlp-mixer_2023,liu_topological_2020}.

\textbf{Parameter Initialization}: Before initializing the weights, parameters such as input size \(X\), motif size \(m\), hidden sizes, sparsity control \(\epsilon\), activation function \(\sigma\), and loss function \(L\) must be defined~\cite{vaswani_attention_2023}.

\textbf{Weights and Bias Initialization}: A random uniform distribution is used for weights, initialized per motif rather than per node. The He function sets bounds, and Erdős–Rényi topology generates sparse masks. The motif size must divide the input size. Biases are initialized similarly.

\vspace{-0.1em} 
\textbf{Listing 2: Network Initialization}
\begin{flushleft}
class MotifBasedSparseNN: \\
\quad Initialize(input\_size, motif\_size, hidden\_sizes, output\_size, init\_network, epsilon, activation\_fn, loss\_fn): \\
\quad\quad Set motif\_size, epsilon, activation\_fn, loss\_fn \\
\quad\quad Set create\_network based on init\_network (uniform / normal) \\
\quad\quad Ensure input\_size is divisible by motif\_size \\
\quad\quad Initialize weights (W) and biases (b) \\
\quad\quad prev\_size = input\_size \\
\quad\quad For each hidden\_size in hidden\_sizes: \\
\quad\quad\quad Ensure hidden\_size is divisible by motif\_size \\
\quad\quad\quad Create and append weight matrix to W using create\_network \\
\quad\quad\quad Create and append bias to b \\
\quad\quad\quad Update prev\_size to hidden\_size \\
\quad\quad Create and append final weight matrix and bias for output\_size
\end{flushleft}

\subsection{Training Process}

\textbf{Forward Propagation}: Nodes are processed in motifs, improving efficiency. Let \(\mathbf{Z}^{(i)} = \mathbf{A}^{(i-1)} \mathbf{W}^{(i)} + \mathbf{b}^{(i)}\) and \(\mathbf{A}^{(i)} = f_{\text{activation}}(\mathbf{Z}^{(i)})\)~\cite{changpinyo_power_2017,bullmore_complex_2009}. Softmax is used at the output layer.

\textbf{Backward Propagation}: The output error is \(\delta^{(L)} = \mathbf{A}^{(L)} - \mathbf{Y}\). Gradients are computed for each layer:

\[
\frac{\partial \mathcal{L}}{\partial \mathbf{W}^{(L)}} = \frac{1}{m} (\mathbf{A}^{(L-1)})^\top \delta^{(L)}, \quad
\frac{\partial \mathcal{L}}{\partial \mathbf{b}^{(L)}} = \frac{1}{m} \sum_{i=1}^{m} \delta^{(L)}
\]

For each motif:
\[
\mathbf{W}_{\text{sub}} = \mathbf{W}^{(i)}[j_{\text{start}}:j_{\text{end}}],
\quad
\delta_{\text{sub}} = (\delta^{(i+1)} \mathbf{W}_{\text{sub}}) \odot f'(\mathbf{Z}_{\text{sub}})
\]
\[
\frac{\partial \mathcal{L}}{\partial \mathbf{W}_{\text{sub}}} = \frac{1}{m} (\mathbf{A}^{(i-1)})^\top \delta_{\text{sub}}, \quad
\delta^{(i)}[j_{\text{start}}:j_{\text{end}}] = \delta_{\text{sub}}
\]

\vspace{0.9em} 
\textbf{Listing 3: MotifBasedSparseNN Training}
\begin{flushleft}
Function backward(X, y\_true, Z\_list, A\_list): \\
\quad m = number of samples \\
\quad Calculate initial delta from loss gradient \\
\quad Calculate dW, db for final layer, Update final weights and biases \\
\quad For i in reverse order of hidden layers: \\
\quad\quad Calculate delta, Initialize dW \\
\quad\quad For each motif in current layer: \\
\quad\quad\quad Calculate sub\_delta and sub\_A \\
\quad\quad\quad Update dW and biases for current motif \\
\quad\quad Update weights and bias for current layer
\end{flushleft}

\subsection{Process of Evolution}

The core of the SET algorithm involves evolution, where weights close to zero are pruned and new weights are assigned~\cite{mocanu_scalable_2018}.

\vspace{0.9em} 
\textbf{Listing 4: Process of Evolution}
\begin{flushleft}
for weight\_matrix in weights: \\
\quad for i in range(weight\_matrix.shape[0]): \\
\quad\quad for j in range(weight\_matrix.shape[1]): \\
\quad\quad\quad if random\_uniform() < epsilon: \\
\quad\quad\quad\quad weight\_matrix[i, j] = 0.0 \quad \# Prune weight \\
\quad\quad\quad weight\_matrix[i, j] += random\_normal() * init\_density \\
return weights
\end{flushleft}

\section{Experiment}
\label{experiment}

This section outlines the experimental process, starting with data preparation, followed by experimental design and evaluation. The aim is to assess the efficiency and accuracy of the motif-based sparse neural network compared to the benchmark model.

\subsection{Data Preparation}

\begin{figure}[h]
    \centering
    \begin{minipage}{0.45\linewidth}
        \centering
        \includegraphics[width=\linewidth]{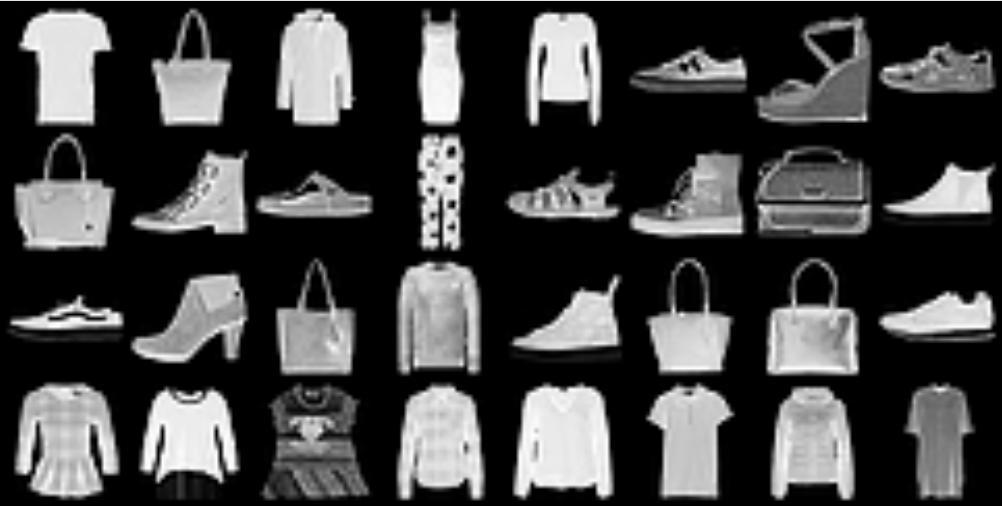}
        \caption{Sample from FMNIST Dataset}
        \label{fig:FMNIST}
    \end{minipage}%
    \hfill
    \begin{minipage}{0.3\linewidth}
        \centering
        \includegraphics[width=\linewidth]{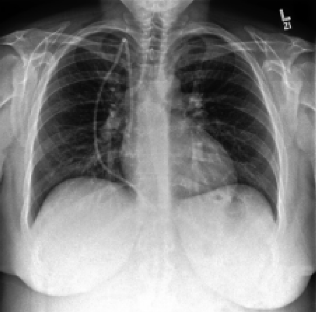}
        \caption{Sample from Lung Dataset}
        \label{fig:Lung}
    \end{minipage}
\end{figure}

In this research, the Fashion MNIST (FMNIST) and Lung datasets are used as benchmarks to evaluate the model’s performance and efficiency~\cite{xiao_fashion-mnist_2017}. The FMNIST dataset (Figure~\ref{fig:FMNIST}) is a widely used benchmark for testing deep learning models. It consists of Zalando’s article images—60{,}000 training samples and 10{,}000 test samples. Each sample is a 28$\times$28 grayscale image with one of 10 categorical labels~\cite{kichler_robustness_2021}.

Images are loaded using TensorFlow’s FMNIST module, pixel values are normalized to the range $[0,1]$, and labels are one-hot encoded. Optionally, data is standardized using \texttt{scikit-learn}'s \texttt{StandardScaler}, and compressed \texttt{.npz} files are generated for efficient reuse.

The Lung dataset includes grayscale X-ray scans with five lung conditions, represented by five labels. After loading, labels are one-hot encoded, and the data is split into training and testing sets (one-third for testing). Normalization using \texttt{StandardScaler} ensures consistent feature scaling.

\begin{table}[H]
    \centering
    \caption{Result of each motif size (FMNIST)}
    \label{num_1}
    \setlength{\tabcolsep}{1.5mm}
    \begin{tabular}{|c|c|c|c|}
        \hline
        Motif Size & Running Time (s) & Accuracy & Avg. Running Time (s) \\
        \hline
        \textbf{1 (SET)} & 25236.2 & 0.7610 & 17.73 \\
        \hline
        \textbf{2}       & 14307.5 & 0.7330 & 9.14  \\
        \hline
        \textbf{4}       & 9209.3  & 0.6920 & 6.74  \\
        \hline
    \end{tabular}
\end{table}

\subsection{Design of the Experiment}

As mentioned, FMNIST and Lung datasets are used to test the accuracy and performance of the motif-based sparse neural network. The SET model~\cite{mocanu_scalable_2018} is used as the benchmark.

A comprehensive setup was implemented, including standardization, one-hot encoding, model initialization, and training. CPU/GPU details were logged, and execution time was recorded. Accuracy and system metrics were saved for detailed analysis. This ensures reproducibility and clarity in performance evaluation~\cite{sun_testing_2019}.

\textbf{FMNIST Design:} The FMNIST dataset has 784 features. The benchmark uses motif size 1; test models use motif sizes 2 and 4. Hidden layers contain 3000 neurons~\cite{wang_rethinking_2024, wang_neural_2018}. A simplified model with two hidden layers (1000 neurons each) is also evaluated.

\textbf{Lung Dataset Design:} The Lung dataset contains 3312 features, divisible by 1, 2, and 4. Models are tested with those motif sizes. A simplified model with two 1000-neuron hidden layers is also implemented.

\textbf{Hyperparameter Setting:} A control variable approach is used. Number of epochs = 300, learning rate = 0.05, and sparsity = 0.1.

\subsubsection*{Comprehensive Score}

To evaluate overall performance, a comprehensive score $S$ is computed based on reductions in running time $R_r$ and accuracy loss $A_r$~\cite{sophia_comprehensive_2024}. Accuracy is weighted more heavily (90\%) than runtime (10\%)~\cite{swink_faster_2006,tan_flexibilityefficiency_2010}:

\begin{equation}
S = 0.1 \times R_r + 0.9 \times (1 - A_r)
\end{equation}
\begin{equation}
R_r = \frac{T_{\text{base}} - T}{T_{\text{base}}}
\end{equation}
\begin{equation}
A_r = \frac{A_{\text{base}} - A}{A_{\text{base}}}
\end{equation}

Where:
\begin{itemize}
    \item $S$: comprehensive score
    \item $R_r$: percentage of running time reduction
    \item $A_r$: percentage of accuracy reduction
    \item $T_{\text{base}}$: benchmark model running time
    \item $T$: running time for the specific motif-size model
    \item $A_{\text{base}}$: benchmark accuracy
    \item $A$: accuracy of the specific motif-size model
\end{itemize}

\section{Results}
\label{sec:results}

This section presents the final results for each dataset and motif size model. The model with the best overall performance per dataset is identified, addressing the research question in Section~\ref{subsec:RQ} by providing exact accuracy and efficiency metrics.

\subsection{Experiment Results}

\subsubsection{FMNIST Results}

In this test, 300 epochs were run with three hidden layers (3000 neurons each). Table~\ref{tab:motif_results} shows that for motif size 1, the total runtime was 25236.2 seconds with 0.761 accuracy. Motif size 2 achieved a runtime of 14307.5 seconds and 0.733 accuracy, reducing runtime by 43.3\% with a 3.7\% drop in accuracy. Motif size 4 achieved a runtime of 9209.3 seconds and 0.692 accuracy, improving efficiency by 73.7\% but with a 9.7\% loss in accuracy.

\begin{figure}[H]
    \centering
    \begin{minipage}{0.48\linewidth}
        \centering
        \includegraphics[width=\linewidth]{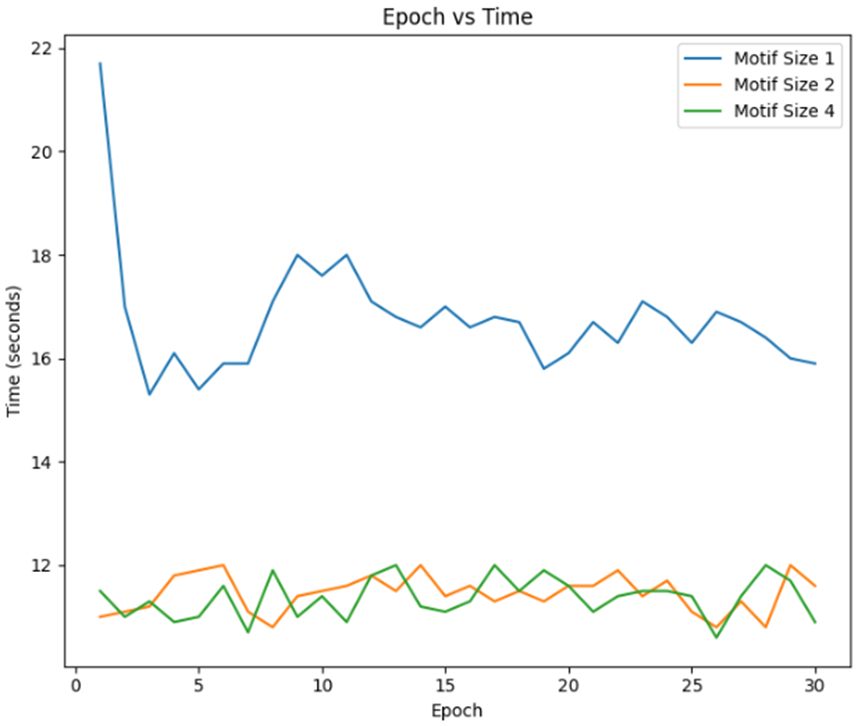}
        \caption{Efficiency of Lung dataset by motif size}
        \label{fig:6}
    \end{minipage}%
    \hfill
    \begin{minipage}{0.48\linewidth}
        \centering
        \includegraphics[width=\linewidth]{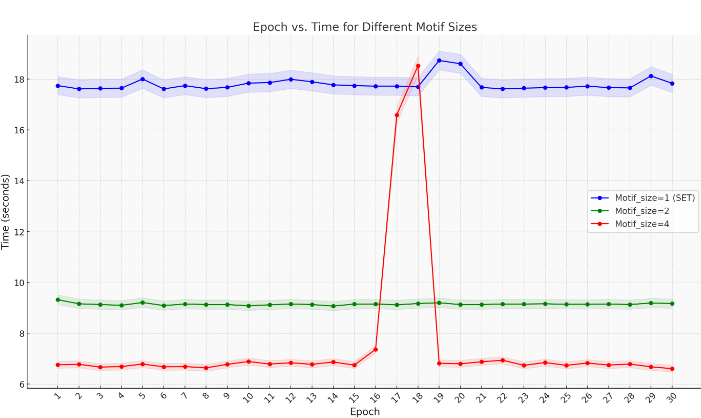}
        \caption{FMNIST Efficiency (First 30 Epochs)}
        \label{fig:5}
    \end{minipage}
\end{figure}

To further evaluate efficiency, a simpler model with two hidden layers (1000 neurons each) was tested. The average running time per epoch over the first 30 epochs is also shown in Figure~\ref{fig:5}.

\begin{table}[t]
\centering
\caption{Motif Size Results (FMNIST)}
\label{tab:motif_results}
\scriptsize
\setlength{\tabcolsep}{1mm}
\begin{tabular}{|c|c|c|c|}
\hline
Motif Size & Time (s) & Accuracy & Avg. Time/Epoch (s) \\
\hline
1 (SET) & 25236.2 & 0.761 & 17.73 \\
2 & 14307.5 & 0.733 & 9.14 \\
4 & 9209.3 & 0.692 & 6.74 \\
\hline
\end{tabular}
\end{table}

Comprehensive scores $S$ were computed as follows:
\vspace{-0.2em} 
\begin{align}
S_1 &= 0.1 \times 0 + 0.9 \times (1 - 0) = 0.9000 \\
S_2 &= 0.1 \times 0.433 + 0.9 \times (1 - 0.037) = 0.9100 \\
S_4 &= 0.1 \times 0.637 + 0.9 \times (1 - 0.097) = 0.8864
\end{align}

Motif size 2 achieved the best overall score (0.9100), outperforming the benchmark model by 1.1\%.

\subsubsection{Lung Dataset Results}

With 300 epochs and three 3000-neuron hidden layers, motif size 1 (benchmark) achieved 0.937 accuracy in 4953.2 seconds. Motif size 2 reduced runtime to 3448.7 seconds with 0.926 accuracy, while motif size 4 further dropped runtime to 3417.3 seconds but with 0.914 accuracy.

Comprehensive score calculations:

\begin{align}
R_{r1} &= 0,\quad A_{r1} = 0,\quad S_1 = 0.9 \\
R_{r2} &= 0.3039,\quad A_{r2} = 0.0117,\quad S_2 = 0.9199 \\
R_{r4} &= 0.3103,\quad A_{r4} = 0.0246,\quad S_4 = 0.9089
\end{align}

Motif size 2 again provides the highest comprehensive score. Efficiency improved by 30.4\% with only a 1.2\% accuracy loss.

\begin{table}[h]
\vspace{-10pt}
\centering
\caption{Dataset Properties}
\label{num_0}
\setlength{\tabcolsep}{2.5mm}
\begin{tabular}{|c|c|c|c|c|}
\hline
Name & Features & Type & Samples & Classes \\
\hline
FMNIST  & 784   & Image       & 70000 & 10 \\
Lung    & 3312  & Microarray  & 203   & 5  \\
\hline
\end{tabular}
\end{table}

\subsection{Result Analysis}

Training observations show that the Lung dataset achieved higher initial accuracy than FMNIST, likely due to its larger feature size (3312 vs. 784) and fewer classes (5 vs. 10)~\cite{xiao_fashion-mnist_2017}.

Across both datasets, motif size 2 models consistently achieved the best trade-off between performance and efficiency. Although motif size 4 offered slightly greater speed, it came at the cost of more significant accuracy loss. The comprehensive scoring function—weighted 0.9 on accuracy and 0.1 on efficiency—confirmed that motif size 2 delivers the optimal overall performance.

\subsection{Trade-off Relationship}

This subsection explores how different efficiency-accuracy weight ratios influence the comprehensive score for each motif size. Figures~\ref{fig:7} and~\ref{fig:8} show that as soon as efficiency becomes a factor (weight $> 0.1$), motif-based models outperform the baseline SET model.

While accuracy remains the dominant factor in most applications, real-world scenarios like autonomous vehicles and embedded systems demand efficient models as well~\cite{han_deep_2016, wu_performance-efficiency_2022, atashgahi_quick_2020}.

\subsection{Application Scenarios}

Motif-based models are especially suitable for use cases requiring both speed and precision~\cite{the_theano_development_team_theano_2016}. Potential application domains include:

\begin{itemize}
    \item \textbf{Mobile Devices}: Resource-constrained environments benefit from efficient inference~\cite{zhang_cambricon-x_2016}.
    \item \textbf{Autonomous Driving}: Real-time decision-making is critical for safety and performance.
    \item \textbf{Financial Trading}: High-frequency trading systems require fast, reliable predictions.
    \item \textbf{Smart Home Systems}: Quick response to sensor input improves user experience and system intelligence.
\end{itemize}

\section{Discussion}
\label{sec:discussion}

This paper proposed the concept of motif-based structural optimization. Building upon the SET-MLP benchmark model with feature engineering, motif-based models were developed and evaluated to identify the configuration with the best performance. According to the results, motif-based models significantly reduce computational cost while incurring a modest drop in accuracy. However, the exact performance of each motif configuration is dataset-dependent. This section discusses these variations and explores the trade-off relationship between efficiency and accuracy, which is central to this research.

\subsection{Result Analysis}

During training and testing, it was observed that the Lung dataset achieved higher precision than FMNIST in the initial training phase (first 30 epochs). This is likely due to the Lung dataset containing 3312 features—nearly four times the 784 features in FMNIST—and having only 5 output classes compared to FMNIST’s 10~\cite{xiao_fashion-mnist_2017}.

Across both datasets, the motif size 2 configuration achieved the best overall performance, combining high efficiency, accuracy, and training stability. The efficiency gain from the baseline model to motif size 2 was more substantial than the gain from motif size 2 to motif size 4. Additionally, the accuracy loss for motif size 2 was smaller than that for motif size 4.

While such trends may suggest that motif size 2 is optimal, a simple observation is insufficient. Therefore, a comprehensive scoring equation was used to evaluate models more rigorously. Given that accuracy typically outweighs efficiency in importance for DNNs, the score weights were set to 0.9 for accuracy and 0.1 for efficiency. However, results indicate that motif-based DNN performance can vary depending on the dataset type, network structure, and use case.

\subsection{Trade-off Relationship}

\begin{figure}[H]
    \centering
    \begin{minipage}{0.48\linewidth}
        \centering
        \includegraphics[width=\linewidth]{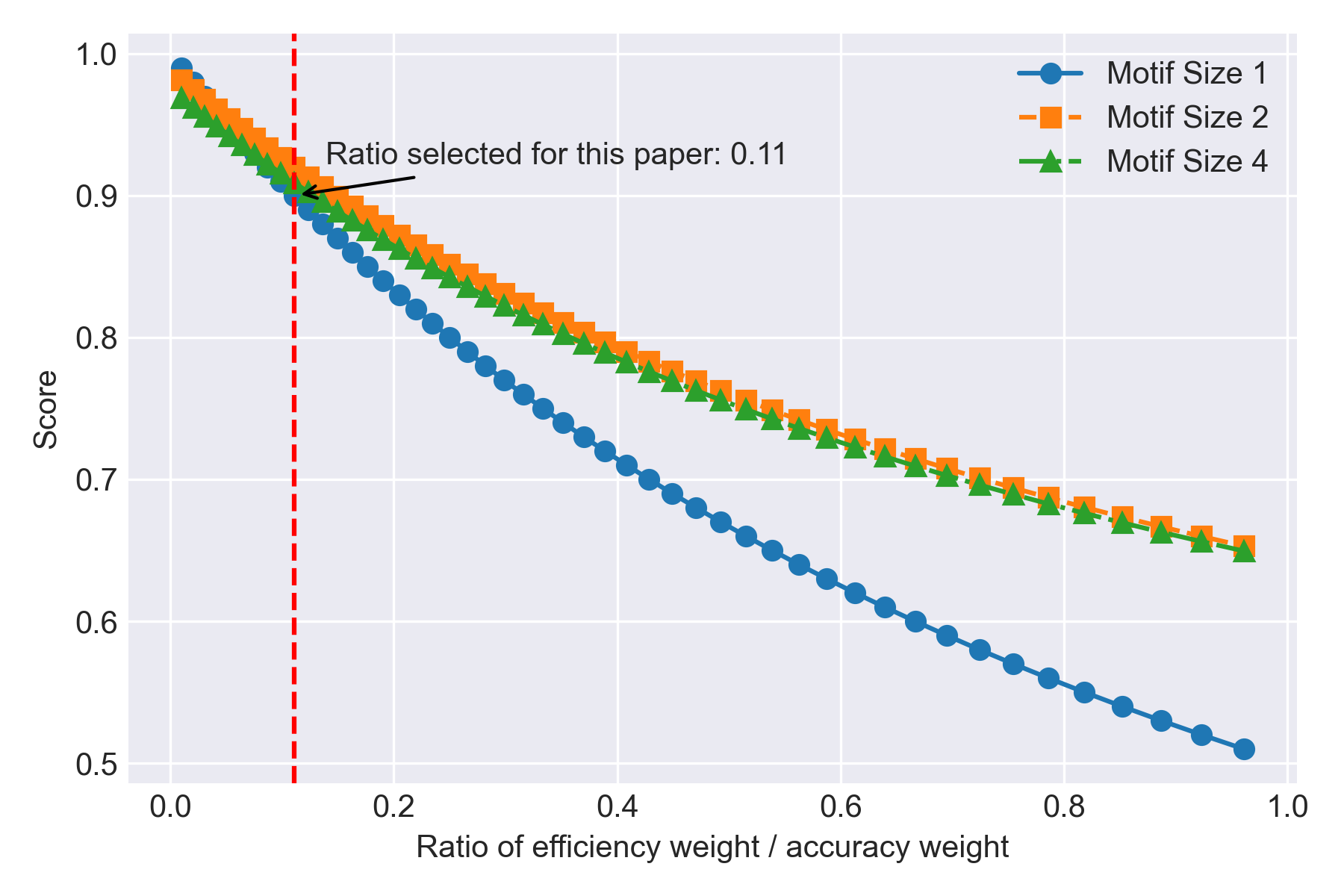}
        \caption{Efficiency vs. Accuracy Weight Ratio and Comprehensive Score (Lung)}
        \label{fig:7}
    \end{minipage}%
    \hfill
    \begin{minipage}{0.48\linewidth}
        \centering
        \includegraphics[width=\linewidth]{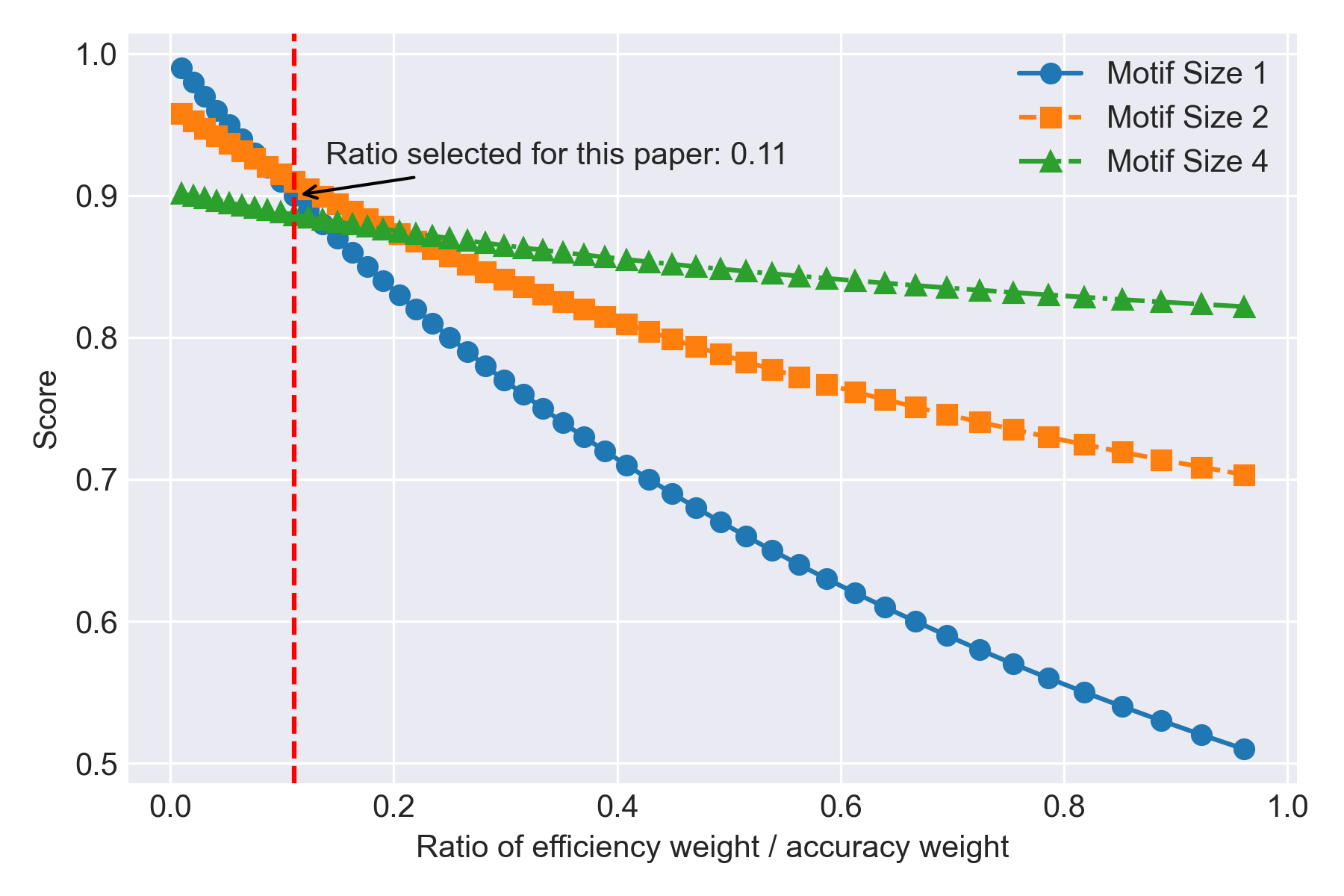}
        \caption{Efficiency vs. Accuracy Weight Ratio and Comprehensive Score (FMNIST)}
        \label{fig:8}
    \end{minipage}
\end{figure}

This subsection explores the trade-off between efficiency and accuracy using the comprehensive score across different weight ratios. Figures~\ref{fig:7} and~\ref{fig:8} illustrate that once efficiency is given even a modest weight (greater than 0.1), motif-based models outperform standard SET models.

While accuracy is usually prioritized in theory, real-world applications often require efficient models as well~\cite{han_deep_2016}. Efficient models provide the responsiveness and resource optimization necessary for use cases such as mobile computing, autonomous systems, and real-time analytics~\cite{wu_performance-efficiency_2022, atashgahi_quick_2020}.

Furthermore, these results demonstrate that overall model performance varies with motif size, dataset properties, and neural network architecture. This confirms the relevance of motif-based approaches for efficiency-critical applications and highlights the need to tailor motif configurations to specific use cases.

\subsection{Application Scenarios}

To explore potential applications of motif-based optimization, this subsection highlights several suitable real-world scenarios. As noted, these models are particularly well-suited for use cases that require low latency and computational efficiency~\cite{the_theano_development_team_theano_2016}. Example domains include:

\begin{itemize}
    \item \textbf{Mobile Devices}: Motif-based models reduce energy and processing demands, enabling efficient inference on battery-limited devices and embedded systems~\cite{zhang_cambricon-x_2016}.
    \item \textbf{Autonomous Driving}: Autonomous vehicles must process sensor data in real-time. Motif-based models can accelerate computation while maintaining prediction accuracy.
    \item \textbf{Financial Trading}: In high-frequency trading, fast and accurate predictions are essential. Motif-based models offer a viable balance.
    \item \textbf{Smart Home Systems}: Responsive behavior is crucial in smart home environments. Motif-based models can efficiently interpret user inputs and environmental signals.
\end{itemize}

\section{Conclusion}
\label{sec:conclusion}

This paper introduced the concept of motif-based structural optimization and demonstrated its application to the SET-MLP feature engineering benchmark model. Through extensive testing, it was shown that motif-based models significantly reduce computational costs while incurring only a slight decrease in accuracy. The analysis revealed that the Lung dataset, with more features and fewer output labels, achieved higher accuracy more quickly compared to the FMNIST dataset.

Among all the tested configurations, the motif size of 2 emerged as the most optimal choice, offering the best balance between efficiency and accuracy. This trade-off was quantified using a comprehensive scoring equation that prioritized accuracy while still valuing efficiency. Notably, for the FMNIST dataset, the motif size 2 model achieved a 43.3\% improvement in efficiency with only a 3.7\% drop in accuracy. For the Lung dataset, it yielded a 30.4\% efficiency gain with just a 1.2\% reduction in accuracy.

In conclusion, motif-based models—particularly with a motif size of 2—demonstrated the best overall performance. The motif-based structural optimization approach is therefore highly effective for scenarios where computational efficiency is critical. The results emphasize the importance of tailoring motif size to the specific dataset and application context to achieve an optimal balance between performance and resource efficiency.

\section{Reflection and Future Work}
\label{sec:Reflection and Future Work}

This study explored the structural optimization of SET using a straightforward motif-based method. Based on experimental results from two benchmark datasets and six different motif size configurations, the findings—summarized in Section~\ref{sec:conclusion}—indicate that a motif size of 2 consistently yields the best overall performance. Moreover, motif-based models generally outperform standard SET models when efficiency is a significant concern.

However, this does not preclude the possibility of more effective structural optimization strategies. For example, incorporating a dynamic motif size selection mechanism during the training process may further enhance model performance. While this study employed a fixed motif size for simplicity and interpretability, future research should examine adaptive strategies.

Additionally, the current results are based on only two datasets. Broader testing across a more diverse set of datasets and application scenarios is essential to validate the generalizability and robustness of motif-based models. These investigations will help determine whether the approach can be effectively applied in varied real-world contexts.

Lastly, this paper introduced a comprehensive score equation to quantify overall model performance by combining efficiency and accuracy. However, there is no universally accepted threshold for balancing these two metrics in machine learning models. Establishing such standards could be a valuable direction for future research in model evaluation and optimization.

\bibliographystyle{plain} % or another style like acm, ieeetr, etc.

\newpage
\onecolumn
% You can choose whether you prefer a single or double column appendix.
% Whatever you choose, you will need to stick to it throughout the appendix.
% For double column style, comment the next line.
\onecolumn

\end{document}